\documentclass{article}

\usepackage[final]{corl_2026} %
\usepackage{amsmath}
\usepackage{amssymb}
\usepackage{float}
\usepackage{placeins}
\usepackage{colortbl}
\definecolor{baselinegray}{gray}{0.93}

\title{MAST-VLA: Skill-Conditioned Adaptation with Trajectory-Affordance Priors for Multi-Skill Robot Manipulation}

\author{
\textbf{
Jiachi Yang$^{1,2}$ \quad
Zhiyi Xu$^{4}$ \quad
Hongquan Zhang$^{1,3}$
} \\
\textbf{
Xin Tan$^{3}$ \quad
Zhizhong Zhang$^{3}$ \quad
Jingyu Gong$^{3,\dagger}$ \quad
Yuan Xie$^{1,3,*}$
} \\
$^{1}$ Shanghai Innovation Institute \quad
$^{2}$ Tianjin University \\
$^{3}$ East China Normal University \quad
$^{4}$ Tongji University
}
\usepackage{graphicx}
\usepackage{booktabs}
\usepackage{multirow}

\begin{document}
\maketitle

\begingroup
\renewcommand{\thefootnote}{\fnsymbol{footnote}}
\footnotetext[2]{Project Leader. \quad $^{*}$ Corresponding author.}
\endgroup
\raggedbottom

\begin{abstract}
Generalist robot manipulation requires a single vision-language-action (VLA) policy to execute diverse skills across scenes and instructions. 
However, current multi-skill VLA policies often leave two aspects of an instruction insufficiently modeled: which primitive skill to execute and where/how to contact and move objects.
We propose \textbf{MAST-VLA} (\textbf{M}ulti-skill \textbf{A}daptation with \textbf{S}kill-conditioned experts and \textbf{T}rajectory-affordance priors), a dual-conditioning framework for robot manipulation.
MAST-VLA introduces skill-conditioned LoRA Mixture-of-Experts (LoRA-MoE) adapters in the LLM backbone, where skill-query tokens and routing biases guide low-rank expert specialization, and converts a predicted affordance heatmap into a patch-level trajectory-affordance spatial prior for the action head.
On the CALVIN ABC$\rightarrow$D long-horizon multi-skill cross-scene generalization setting, MAST-VLA achieves state-of-the-art performance, reaching an Avg. Len of 4.65 over five-subtask evaluation sequences. 
Real-robot multi-skill sequence experiments further validate its deployability under real perception and control noise.
\end{abstract}

\section{Introduction}

Vision-language-action models (VLAs) provide a unified interface for language-conditioned robot manipulation by connecting visual perception, language understanding, and action generation~\citep{rt2, openvla, pi0, pi05}. 
Recent VLA policies show that pretrained multimodal models can be adapted to robot control across diverse objects, scenes, tasks, and embodiments~\citep{roboflamingo, octo, openvla_oft, gr00t_n1}. 
This capability makes VLAs a promising foundation for generalist robot manipulation, where a single policy is expected to reuse visual-language knowledge while executing and composing multiple manipulation skills~\citep{rt1, openx, bc_z}.

Despite this progress, multi-skill long-horizon manipulation remains challenging because language-conditioned policies must maintain stable execution across sequential subtasks and skill transitions~\citep{calvin, hulc, hulc_plus, spil}. 
A manipulation instruction requires the policy to infer the intended primitive operation, such as grasping, pushing, rotating, or stacking, while also grounding the contact regions, motion paths, and object interactions required for execution~\citep{cliport, peract, rvt, where2act, vat_mart}. 
These two factors are referred to in this work as \emph{primitive-skill semantics} and \emph{object-interaction semantics}. 
When either factor is insufficiently modeled, errors may accumulate across consecutive subtasks, leading to unstable execution under long-horizon skill transitions~\citep{calvin, hulc_plus, lumos}.

Existing VLA policies often leave these two forms of grounding implicit in language instructions,
visual observations, and low-dimensional action supervision~\citep{openvla, openvla_oft, vla_adapter, x_vla}. 
For skill-level grounding, shared adaptation parameters or task-level lightweight modules may not explicitly align expert capacity with primitive manipulation skills, making it difficult to separate skill-specific action patterns and control preferences within a shared policy~\citep{lora, mixture_of_lora_experts, vla_adapter, coral}. 
For interaction-level grounding, low-dimensional action supervision provides only indirect cues about contact regions, motion trends, and task-relevant geometry, whereas prior manipulation methods have shown the importance of explicit spatial, geometric, or affordance-level signals~\citep{transporter, cliport, peract, rvt, affordance_human_video, robopoint}. 
Thus, multi-skill VLA policies require mechanisms that preserve shared visual-language knowledge while explicitly conditioning adaptation on primitive skills and strengthening action-relevant spatial supervision.

To address these limitations, we propose \textbf{MAST-VLA}, a dual-conditioning framework for multi-skill robot manipulation.
MAST-VLA combines skill-conditioned LoRA-MoE specialization in the LLM backbone with a trajectory-affordance spatial prior in the action head.
In the LLM backbone, skill-query tokens and routing biases condition low-rank expert specialization; in the action head, trajectory-derived affordance heatmaps are converted into patch-level attention biases for action decoding.
Together, these two paths reduce cross-skill interference and strengthen interaction-aware action generation.

The contributions are summarized as follows:

1) A dual-conditioning VLA framework is proposed for multi-skill robot manipulation, jointly modeling primitive-skill structure and interaction-relevant spatial grounding within a shared policy;

2) Skill-conditioned LoRA-MoE specialization and trajectory-affordance action guidance are introduced. The former uses skill-query tokens and routing biases to condition low-rank expert specialization in the LLM backbone, while the latter converts future gripper motion and scene geometry into heatmap supervision and patch-level spatial bias in the action head;

3) MAST-VLA achieves state-of-the-art performance on CALVIN ABC$\rightarrow$D with an Avg. Len of 4.65 and improves Success@5 from 80.0\% to 84.7\%. Real-robot experiments, ablations, and visualizations further validate sequential robustness and the complementarity of the two paths.

\begin{figure}[t]
    \centering
    \includegraphics[width=\textwidth]{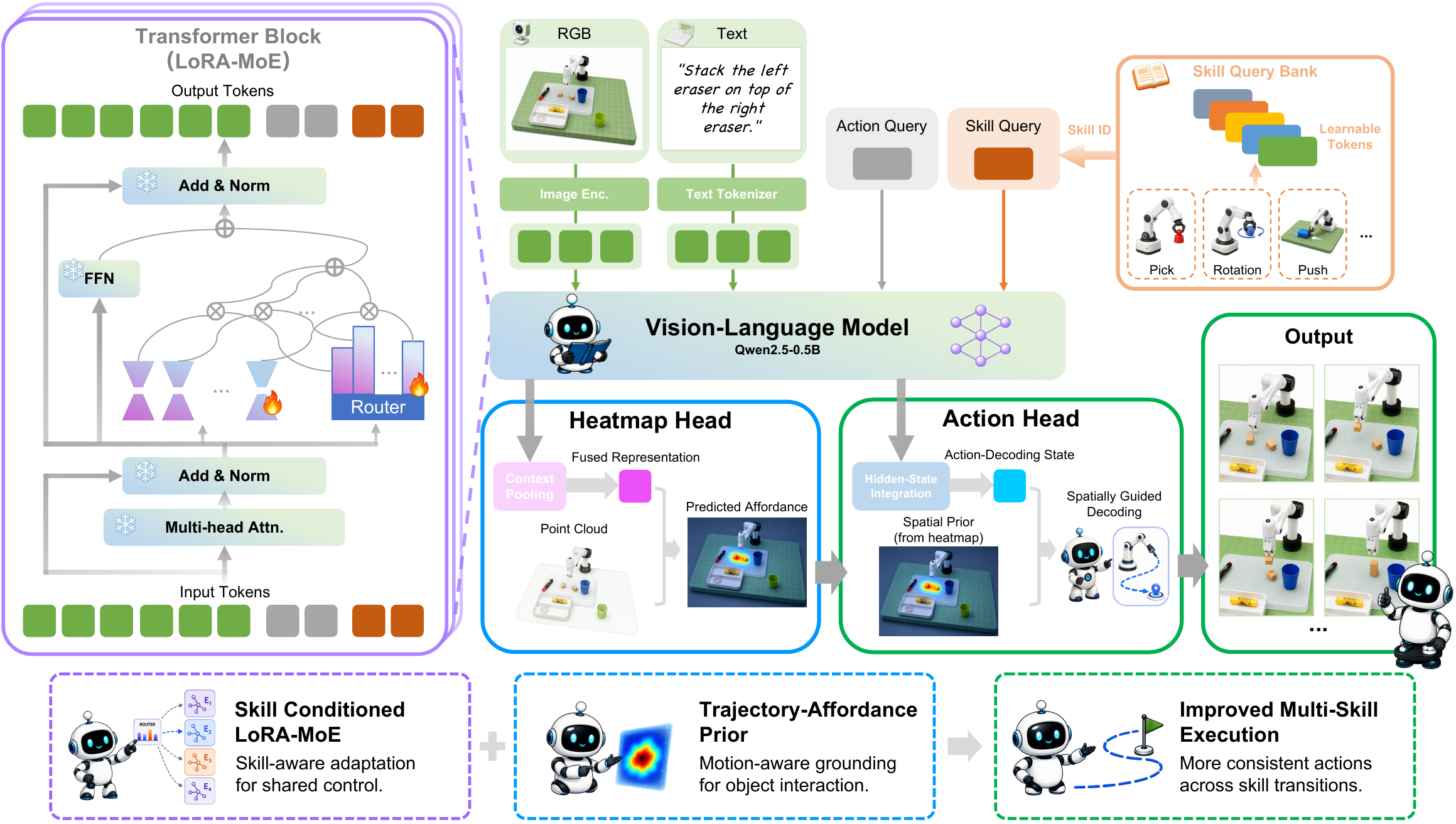}
    \caption{
    \textbf{Overview of MAST-VLA.}
    MAST-VLA introduces two complementary conditioning paths for multi-skill VLA policies: skill-conditioned LoRA-MoE adaptation in the LLM backbone and a trajectory-affordance spatial prior in the action head. The expanded LLM-block view shows FFN-side LoRA-MoE routing with skill-conditioned routing bias, while the action-head view shows how the predicted affordance heatmap is converted into a patch-level spatial prior for action generation.
    }
    \label{fig:method_overview}
\end{figure}

\section{Related Work}
\label{sec:related_work}

\paragraph{Generalist VLA Policies for Robot Manipulation.}
Generalist robot policies and vision-language-action models aim to map visual observations and language instructions to robot actions across tasks, scenes, and embodiments~\citep{rt1, rt2, openx, octo, pi0}.
Recent VLA policies transfer pretrained visual-language knowledge to manipulation and support efficient adaptation to downstream robot control~\citep{palm_e, roboflamingo, openvla, openvla_oft, vla_adapter}.
Action modeling has also evolved from autoregressive action tokens to action chunking, diffusion policies, and fast action decoding~\citep{act, diffusion_policy, fast}.
These advances provide strong generalist priors, but multi-skill adaptation still often leaves primitive-skill separation and interaction grounding implicit in language-action supervision.
MAST-VLA treats multi-skill VLA adaptation as a dual-grounding problem: aligning adaptation capacity with primitive skill structure and grounding action decoding in trajectory-derived interaction geometry.

\paragraph{Modular Adaptation and Skill Specialization.}
Parameter-efficient adaptation methods such as adapters, prompt tuning, and LoRA specialize pretrained models with a small number of additional parameters~\citep{adapter, prompt_tuning, lora}.
For robot policies, lightweight adaptation modules, action-head adaptation, and task-level LoRA experts improve efficient specialization and can reduce multi-task interference~\citep{openvla_oft, vla_adapter, t2s, coral}.
MoE and LoRA-MoE methods further increase adaptation capacity through sparse expert routing~\citep{sparsely_gated_moe, switch_transformer, mixture_of_lora_experts}.
However, existing VLA adaptation methods are mainly single-path or task-level, either sharing the same lightweight update across skills or allocating experts by tasks/domains rather than primitive operations.
MAST-VLA uses the instruction-derived primitive skill condition to modulate both input-side skill queries and router-side biases, thereby aligning expertized adaptation capacity with primitive robot skills within a shared VLA backbone.

\paragraph{Spatial and Affordance Grounding for Manipulation.}
Robot manipulation requires spatial reasoning about contact regions, motion paths, and 3D object relations.
Prior work improves spatial awareness through transport-based representations, voxel or multi-view policies, and 3D-aware manipulation models~\citep{transporter, cliport, peract, rvt, dp3, robmrag}.
Recent VLA methods also incorporate 3D representations, spatial projections, or geometry-aware visual features into robot control~\citep{pointvla, geovla, spatialvla, falcon_vla}.
Another line of work provides explicit spatial cues through affordance learning, interaction-region prediction, or trajectory-conditioned visual prompts~\citep{where2act, vat_mart, affordance_human_video, robopoint, tracevla}.
Unlike methods that mainly use spatial cues as observation representations, target regions, or input prompts, MAST-VLA converts demonstration trajectories into a dense action-decoding prior and injects the predicted heatmap into action-head attention for multi-skill action generation.
\section{Methodology}
\label{sec:method}

\paragraph{Problem setup.}
This work considers sequential multi-skill robot manipulation, where a shared vision-language-action policy executes successive primitive instructions under an evolving scene state.
Given the current observation $\mathbf{o}_t$, language instruction $\mathbf{l}$, and coarse primitive skill condition $s_t$, the model predicts a future $H$-step action sequence $\mathbf{a}_{t:t+H-1}$:
\begin{equation}
\label{eq:policy}
\pi_\theta:\;(\mathbf{o}_t,\mathbf{l},s_t)\mapsto \hat{\mathbf{a}}_{t:t+H-1}.
\end{equation}
Here, $\mathbf{o}_t$ contains visual observations and robot proprioceptive states, and $s_t$ denotes a coarse primitive skill condition associated with the current instruction.
Fine-grained manipulation commands are grouped into a compact primitive skill taxonomy, such as rotation, sliding/pushing, lifting/picking, placing/storing, stacking, and switching.
For each step, the language instruction is matched to instruction variants with normalized lexical and fuzzy matching, and the matched operation is grouped into this taxonomy to obtain the language-derived primitive skill condition $s_t$.
This condition only indexes skill-query and routing-bias banks; the original language instruction and action target remain unchanged.
\paragraph{Framework overview.}
As shown in Figure~\ref{fig:method_overview}, MAST-VLA contains two coupled conditioning paths for multi-skill VLA policy learning.
The first path performs skill-conditioned LoRA-MoE adaptation in the LLM backbone, where the instruction-derived primitive skill condition modulates token representations and expert routing to form skill-aware adaptation paths within a shared policy (Sec.~\ref{sec:skill_lora_moe}).
The second path introduces a trajectory-affordance spatial prior in the action head, where demonstration-derived gripper motion and scene geometry provide affordance-heatmap supervision and patch-level attention guidance for action decoding (Sec.~\ref{sec:heatmap_supervision}).
The full model is trained with action regression, affordance-heatmap supervision, and MoE routing regularization (Sec.~\ref{sec:training_objective}), coupling skill-level specialization with interaction-aware spatial grounding.

\subsection{Skill-Conditioned LoRA-MoE Adaptation}
\label{sec:skill_lora_moe}

Skill-conditioned LoRA-MoE adaptation is implemented by replacing selected FFN/MLP sublayers in the LLM backbone with LoRA-MoE adapters.
Each LoRA-MoE adapter contains multiple low-rank experts and uses top-$k$ routing to select a small number of expert paths for each token.

\paragraph{Skill-specific condition banks.}
Assume the primitive skill taxonomy contains $N_s$ coarse skill categories, each represented by $N_q$ learnable skill-query tokens, and the hidden dimension is $d$.
An input-side skill query bank and a layer-wise routing bias bank are maintained for LoRA-MoE adapters:
\begin{equation}
\label{eq:skill_condition_banks}
\mathbf{E}_{\mathrm{query}}\in\mathbb{R}^{N_s\times N_q\times d},
\qquad
\mathbf{E}_{\mathrm{route}}^{(\ell)}\in\mathbb{R}^{N_s\times N_e}.
\end{equation}
Given the primitive skill condition $s_t$, the corresponding skill queries $\mathbf{Q}_{s_t}=\mathbf{E}_{\mathrm{query}}[s_t]$ are retrieved and concatenated with visual tokens, language tokens, and action queries:
\begin{equation}
\label{eq:skill_conditioned_tokens}
\mathbf{X}_t=[\mathbf{V}_t;\mathbf{L}_t;\mathbf{A};\mathbf{Q}_{s_t}].
\end{equation}
For the LoRA-MoE adapter in the $\ell$-th Transformer block, the model also retrieves the routing bias $\mathbf{b}_{s_t}^{(\ell)}=\mathbf{E}_{\mathrm{route}}^{(\ell)}[s_t]$ to modulate expert routing.
Importantly, $s_t$ is not inserted into the language stream as an additional discrete token and is not used as an action target. It only indexes learnable continuous conditioning parameters, while the original language instruction and action supervision remain unchanged.

\paragraph{LoRA-MoE adapter with skill-conditioned routing.}
For the LoRA-MoE adapter in the $\ell$-th Transformer block, given a token hidden state $\mathbf{h}$, the router computes expert weights from content features and the skill-conditioned routing bias:
\begin{equation}
\label{eq:skill_conditioned_router}
\mathbf{p}
=
\operatorname{softmax}\!\left(
\mathbf{W}_{\mathrm{gate}}\mathbf{h}
+
\mathbf{b}_{s}^{(\ell)}
\right).
\end{equation}
It then selects the top-$k$ experts to form the set $\mathcal{T}$, renormalizes the routing weights within this set, and computes the LoRA-MoE output update:
\begin{equation}
\label{eq:lora_moe_update}
\Delta\mathbf{h}
=
\sum_{i\in\mathcal{T}}
\bar{p}_i\,\mathbf{B}_i\mathbf{A}_i\mathbf{h},
\qquad
\mathcal{T}=\operatorname{TopK}(\mathbf{p},k).
\end{equation}
Here, $\mathbf{A}_i$ and $\mathbf{B}_i$ are the low-rank adaptation matrices of the $i$-th LoRA expert, and $\bar{p}_i$ denotes the routing weight after renormalization within the selected top-$k$ experts.
Since the routing bias only shifts content-based routing, expert selection remains context-adaptive rather than being hard-bound to a skill, forming soft skill-conditioned adaptation paths in the shared backbone.

\subsection{Trajectory-Affordance Spatial Prior}
\label{sec:heatmap_supervision}

\paragraph{Affordance heatmap target construction.}
Low-dimensional action supervision only constrains the final action chunk and provides limited direct supervision for task-relevant spatial structures in visual representations.
An affordance heatmap target is therefore constructed automatically from the 3D distance between the current scene surface and the future gripper trajectory.
Let $\mathbf{x}_u$ be the 3D surface point back-projected from pixel $u$, and let $\{\mathbf{g}_{t},\ldots,\mathbf{g}_{t+H-1}\}$ denote the future $H$-step gripper positions. We define
\begin{equation}
\label{eq:trajectory_affordance_target}
\begin{aligned}
d_u^2
&=
\min_{\tau\in\{t,\ldots,t+H-1\}}
\|\mathbf{x}_u-\mathbf{g}_{\tau}\|_2^2,\\
H^*(u)
&=
\exp\!\left(-\frac{d_u^2}{2\sigma^2}\right)
\mathbb{I}(d_u\le r_{\max}).
\end{aligned}
\end{equation}
Here, $\sigma$ controls the spatial spread of the response, and $r_{\max}$ truncates background regions far from the trajectory.
Since the future gripper trajectory is taken from the expert demonstration associated with the current instruction, $H^*$ projects the motion tendency of skill execution back to the current observation space, providing dense geometric supervision without additional manual annotation.
Additional details on RGB-D preprocessing, point-map construction, and heatmap loss weighting are provided in Appendix~\ref{app:heatmap_target_details}.

\paragraph{Dense trajectory-affordance supervision.}
The affordance-heatmap head predicts $\hat{H}$ from intermediate VLA representations and an RGB-D-derived point map of the current scene, and is supervised by the trajectory-affordance target $H^*$.
As a dense geometric target over visible scene surfaces near future gripper motion, $H^*$ encourages the predicted heatmap to capture operation-relevant contact regions and motion trends.
Rather than being used only as an auxiliary output, the predicted heatmap is further converted into a spatial prior for action decoding.

\paragraph{Affordance-heatmap-guided action decoding.}
To make the predicted affordance heatmap directly influence action decoding rather than serve only as auxiliary supervision, it is normalized and downsampled into a spatial prior $\mathbf{M}\in\mathbb{R}^{N_v}$ aligned with visual patch tokens.
In the action head, $\mathbf{M}$ is injected as an additive bias into the attention logits associated with visual tokens:
\begin{equation}
\label{eq:heatmap_guided_attention}
\mathbf{A}
=
\operatorname{softmax}\!\left(
\frac{\mathbf{Q}\mathbf{K}_{\mathrm{vis}}^{\top}}{\sqrt{d_h}}
+
\tau \mathbf{M}
\right),
\end{equation}
where $N_v$ is the number of visual tokens, $d_h$ is the attention head dimension, and $\tau$ controls the strength of the spatial prior.
The patch-level spatial prior $\mathbf{M}$ only changes the relative weights of different visual positions in attention and does not mask out non-highlighted regions. Therefore, the action head can still access the full visual context while assigning higher relative attention to interaction regions relevant to the current skill.

\subsection{Training Objective}
\label{sec:training_objective}

MAST-VLA is optimized under a joint objective comprising action-chunk regression, affordance-heatmap supervision, and MoE routing regularization:
\begin{equation}
\label{eq:training_objective}
\mathcal{L}
=
\left\|
\hat{\mathbf{a}}_{t:t+H-1}
-
\mathbf{a}_{t:t+H-1}
\right\|_1
+
\lambda_{\mathrm{hm}}\mathcal{L}_{\mathrm{hm}}
+
\lambda_{\mathrm{moe}}\mathcal{L}_{\mathrm{moe}}.
\end{equation}
The predicted future action chunk is supervised by an $\ell_1$ regression loss.
The affordance-heatmap supervision term $\mathcal{L}_{\mathrm{hm}}$ is instantiated as a weighted binary cross-entropy (WBCE) loss between the predicted affordance heatmap $\hat{H}$ and the trajectory-affordance target $H^*$.
The routing regularization term $\mathcal{L}_{\mathrm{moe}}$ denotes the standard router auxiliary loss, which stabilizes expert assignment and discourages routing collapse~\citep{switch_transformer}.
The coefficients $\lambda_{\mathrm{hm}}$ and $\lambda_{\mathrm{moe}}$ weight the affordance-heatmap supervision and routing regularization terms, respectively.
\section{Experiments}
\label{sec:experiments}

\subsection{Experimental Setup}
\label{sec:experimental_setup}

MAST-VLA is evaluated in both simulation and real-world robot settings.

\noindent\textbf{Simulation.}
The CALVIN ABC$\rightarrow$D benchmark~\citep{calvin} is used to evaluate long-horizon multi-skill generalization, with the policy trained on scenes A--C and tested on the unseen scene D, which differs in visual textures and the spatial arrangement of static scene elements.
Following the standard five-subtask protocol, each evaluation sequence executes five consecutive subtasks without resetting the scene between subtasks.
Success@1--5 and Avg. Len are reported, where Success@k denotes the percentage of evaluation sequences in which at least $k$ consecutive subtasks are completed and Avg. Len denotes the mean completed sequence length.
Detailed protocol and metric definitions are provided in Appendix~\ref{app:calvin_protocol}.

\noindent\textbf{Real-world robot.}
Physical multi-skill manipulation is further evaluated under perception and control noise using two tabletop scenes with length-3 primitive-subtask sequences.
Detailed task sequences and success criteria are provided in Appendix~\ref{app:real_robot_protocol}.
Unless otherwise specified, VLA-Adapter-Pro is used as the direct baseline.

\subsection[Main Results on CALVIN ABC-D]{Main Results on CALVIN ABC$\rightarrow$D}
\label{sec:calvin_main_results}

\begin{table}[!h]
\centering
\caption{Long-horizon robotic manipulation evaluation on CALVIN ABC$\rightarrow$D.}
\label{tab:calvin_main}
\small
\setlength{\tabcolsep}{7pt}
\begin{tabular}{lcccccc}
\toprule
\textbf{Method} &
\multicolumn{5}{c}{\textbf{Tasks Completed in a Row (\%)}} &
\textbf{Avg. Len}$\uparrow$ \\
\cmidrule(lr){2-6}
& \textbf{1} & \textbf{2} & \textbf{3} & \textbf{4} & \textbf{5} & \\
\midrule
OpenVLA~\citep{openvla}
& 91.3 & 77.8 & 62.0 & 52.1 & 43.5 & 3.27 \\
UniVLA~\citep{univla}
& 95.5 & 85.8 & 75.4 & 66.9 & 56.5 & 3.80 \\
PointMapPolicy~\citep{pointmappolicy}
& 96.1 & 88.6 & 80.5 & 72.3 & 63.6 & 4.01 \\
EL3DD~\citep{el3dd}
& 96.6 & 90.0 & 82.4 & 74.2 & 66.2 & 4.09 \\
OpenVLA-OFT~\citep{openvla_oft}
& 96.3 & 89.1 & 82.4 & 75.8 & 66.5 & 4.10 \\
Seer-Large~\citep{seer}
& 96.3 & 91.6 & 86.1 & 80.3 & 74.0 & 4.28 \\
VPP~\citep{vpp}
& 95.7 & 91.2 & 86.3 & 81.0 & 75.0 & 4.33 \\
FALCON~\citep{falcon_vla}
& 98.4 & 94.5 & 88.6 & 82.5 & 75.5 & 4.40 \\
VLA-Adapter-Pro~\citep{vla_adapter}
& 98.5 & 95.0 & 90.5 & 85.3 & 80.0 & 4.50 \\
FLOWER~\citep{flower_vla}
& 99.3 & 95.9 & 90.5 & 84.8 & 77.5 & 4.54 \\
\midrule
\textbf{MAST-VLA (Ours)}
& \textbf{99.4} & \textbf{97.4} & \textbf{94.2} & \textbf{89.2} & \textbf{84.7} & \textbf{4.65} \\
\bottomrule
\end{tabular}
\vspace{-4pt}
\end{table}

Table~\ref{tab:calvin_main} compares MAST-VLA with representative prior methods on CALVIN ABC$\rightarrow$D.
MAST-VLA achieves the best Avg. Len of 4.65, indicating stronger long-horizon multi-skill generalization in the unseen scene D.

The improvement is more pronounced in the later stages of the five-subtask sequences.
Compared with the direct baseline, MAST-VLA improves Success@4 and Success@5 from 85.3\% and 80.0\% to 89.2\% and 84.7\%, respectively.
Since early-stage success rates are already close to saturation, the larger gains at Success@4--5 suggest improved robustness when skill-switching errors and spatial-grounding errors accumulate over consecutive subtasks.

\subsection{Real-World Robot Experiments}
\label{sec:real_robot}

\begin{figure}[t]
\centering
\vspace{-10pt}
\includegraphics[width=0.95\linewidth]{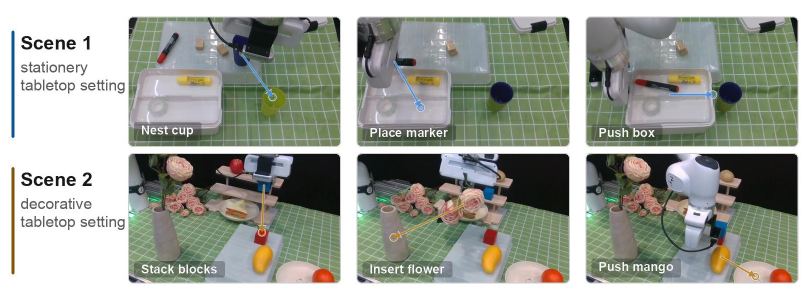}
\vspace{-5pt}
\caption[Representative real-robot evaluation tasks]{
\textbf{Representative real-robot evaluation tasks.}
Representative test-time primitive subtasks are shown for two tabletop scenes. The examples cover diverse interaction modes, including nesting, pick-and-place, pushing, stacking, and insertion.
}
\label{fig:real_robot_tasks}
\end{figure}

\begin{figure}[t]
\centering
\vspace{-4pt}
\includegraphics[width=\linewidth]{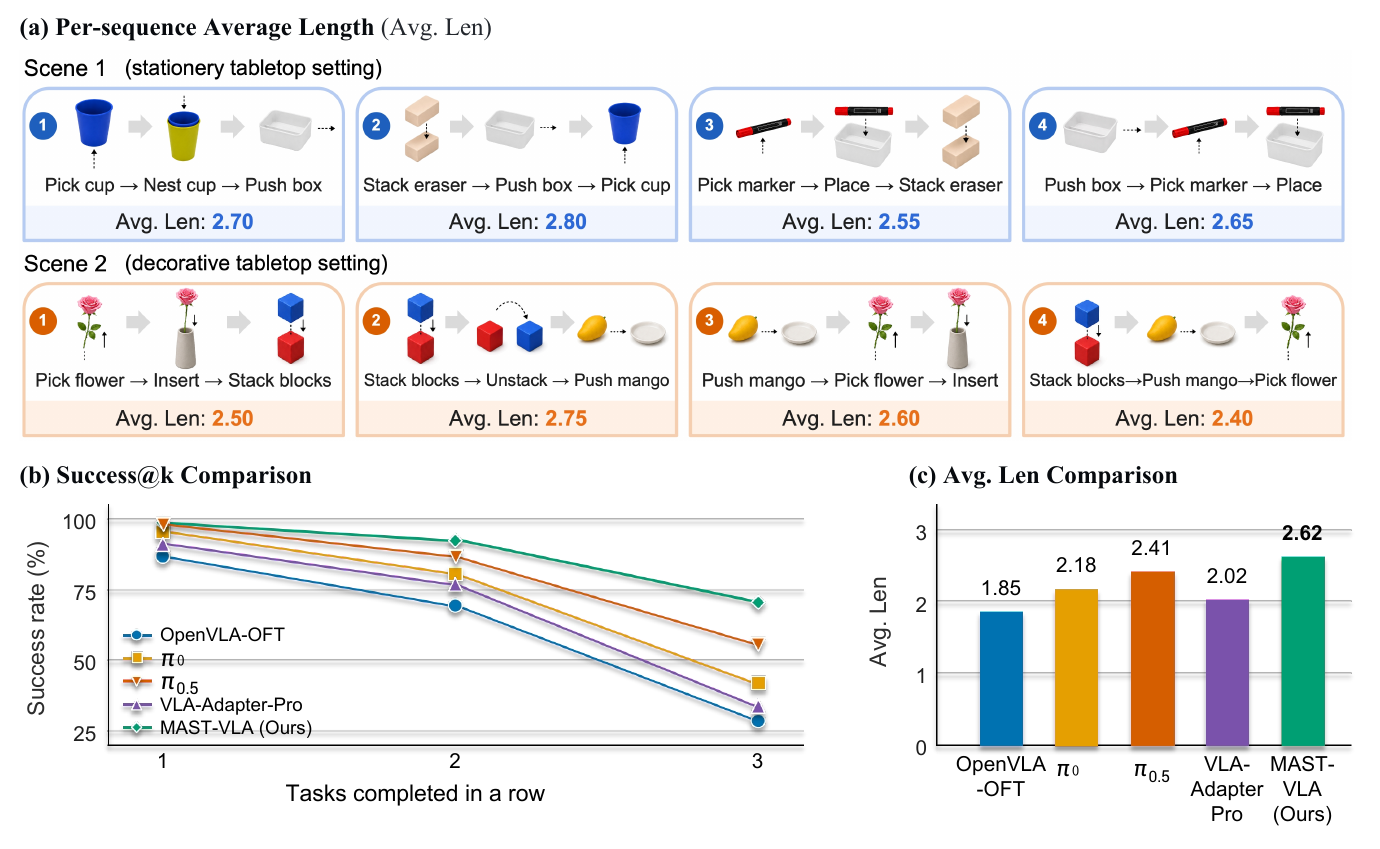}
\vspace{-18pt}
\caption[Real-robot sequence-completion results]{
\textbf{Real-robot sequence-completion results.}
(a) Per-sequence Avg. Len across eight length-3 sequences in two tabletop scenes.
(b) Success@1--3 comparison over consecutive primitive completion.
(c) Overall Avg. Len comparison across all real-robot episodes.
}
\label{fig:real_robot_results}
\end{figure}

Real-robot experiments are conducted on a Franka Research 3 platform equipped with the Franka gripper and two Intel RealSense D435i cameras.
The benchmark contains length-3 primitive-task sequences in two tabletop scenes; instructions are issued sequentially, with the next primitive given only after the current one is completed.
Representative test-time primitive subtasks are shown in Figure~\ref{fig:real_robot_tasks}.

For real-robot training, each primitive task contains approximately 80--100 demonstrations with moderate object-position variations and paraphrased language commands.
All baselines are fine-tuned on the same demonstrations, language commands, state/action targets, and RGB observations using their official pretrained weights and default fine-tuning settings; MAST-VLA additionally uses static-view depth for the trajectory-affordance branch.
Detailed hardware configuration, concrete evaluation sequences, and success criteria are provided in Appendix~\ref{app:real_robot_protocol}.

Success@1--3 and Avg. Len are reported over all real-robot episodes; an episode failing at the $k$-th primitive has completed length $k-1$.
Quantitative sequence-completion results are summarized in Figure~\ref{fig:real_robot_results}.
MAST-VLA achieves higher per-sequence Avg. Len and stronger overall sequence-completion performance than the baselines, suggesting improved robustness to accumulated perception and control errors across consecutive primitives.
Detailed per-sequence Success@1--3 and Avg. Len results are provided in Appendix~\ref{app:real_robot_results}.

\FloatBarrier
\subsection{Ablation and Analysis}
\label{sec:ablation_analysis}

\begin{table}[!htbp]
\centering
\caption{
\textbf{Ablation on CALVIN ABC$\rightarrow$D.}
Query: skill-query tokens.
LoRA-MoE: skill-conditioned low-rank experts.
Aff. Heatmap: trajectory-affordance spatial prior.
Skill Cond.: matched uses the instruction-derived primitive skill condition; mismatched replaces it with a random non-matching condition at test time.
}
\label{tab:ablation}
\small
\setlength{\tabcolsep}{3.2pt}
\renewcommand{\arraystretch}{1.05}
\resizebox{\textwidth}{!}{
\begin{tabular}{cccccccccc}
\toprule
\textbf{Query} &
\textbf{LoRA-MoE} &
\textbf{Aff. Heatmap} &
\textbf{Skill Cond.} &
\multicolumn{5}{c}{\textbf{Tasks Completed in a Row (\%)}} &
\textbf{Avg. Len}$\uparrow$ \\
\cmidrule(lr){5-9}
 & & & & \textbf{1} & \textbf{2} & \textbf{3} & \textbf{4} & \textbf{5} & \\
\midrule
\rowcolor{baselinegray}
 &  &  & -- 
& 98.5 & 95.0 & 90.5 & 85.3 & 80.0 & 4.50 \\

\checkmark &  &  & matched
& 98.7 & 95.6 & 91.3 & 86.4 & 81.1 & 4.53 \\

\checkmark & \checkmark &  & matched
& 98.8 & 96.3 & 92.8 & 88.0 & 83.1 & 4.59 \\

 &  & \checkmark & -- 
& 99.0 & 96.7 & 93.4 & 88.6 & 83.7 & 4.61 \\

\checkmark & \checkmark & \checkmark & matched
& \textbf{99.4} & \textbf{97.4} & \textbf{94.2} & \textbf{89.2} & \textbf{84.7} & \textbf{4.65} \\

\checkmark & \checkmark & \checkmark & mismatched
& 98.8 & 96.2 & 92.7 & 86.6 & 80.9 & 4.55 \\
\bottomrule
\end{tabular}
}
\vspace{-2pt}
\end{table}
\begin{figure}[!htbp]
\centering
\vspace{-3pt}
\includegraphics[width=0.99\linewidth,height=1.00\textheight,keepaspectratio]{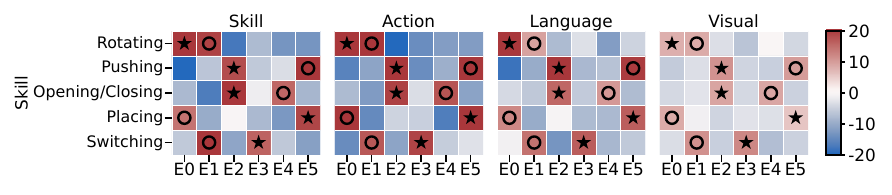}
\vspace{-4pt}
\caption[Skill-conditioned LoRA-MoE routing signatures]{
\textbf{Skill-conditioned Top-2 LoRA-MoE routing signatures.}
Expert-usage shifts across primitive skills and token groups are measured in percentage points (pp) relative to average usage.
Stars/circles denote the experts with the highest/second-highest average routing usage for each skill; they indicate learned routing preferences rather than fixed skill-to-expert assignments.
}
\label{fig:moe_routing}
\end{figure}

\paragraph{Quantitative ablations.}
Table~\ref{tab:ablation} evaluates the contribution of each conditioning path and the reliability of the primitive skill condition.
Skill-query tokens alone provide limited gains, suggesting that input-side skill cues offer useful but relatively weak conditioning.
The larger gain obtained after adding LoRA-MoE suggests that skill conditioning is more effective when paired with modular adaptation capacity, where expert usage can be biased toward skill-specialized action patterns.
The trajectory-affordance spatial prior further improves long-horizon completion by guiding action decoding toward future interaction regions.
The full model performs best among the ablated variants.
Using the same trained checkpoint, replacing the matched skill condition with a random non-matching one at test time reduces Avg. Len from 4.65 to 4.55, indicating that the instruction-derived skill condition provides useful alignment for skill queries and routing biases rather than acting as an unused identifier.

\paragraph{Routing analysis.}
Figure~\ref{fig:moe_routing} visualizes the Top-2 LoRA-MoE routing signatures across skills and token groups.
Different skills exhibit distinct primary and secondary expert preferences, showing that the skill-conditioned bias induces skill-dependent expert specialization.
At the same time, expert usage is shifted rather than fixed and remains coherent across token groups, suggesting soft skill-conditioned routing that preserves content-dependent expert selection.
Together with the skill-query ablation in Table~\ref{tab:ablation}, this provides evidence for primitive-skill semantics through skill-conditioned token representations and expert routing.
\begin{figure}[!t]
\centering
\vspace{-4pt}
\includegraphics[width=1.00\linewidth]{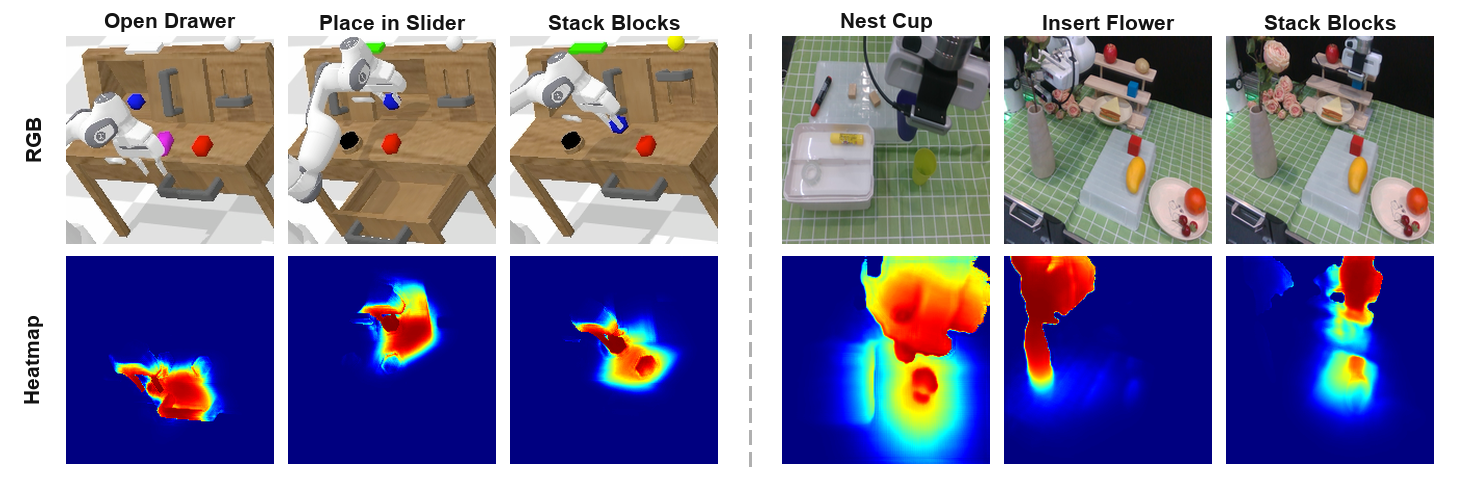}
\vspace{-10pt}
\caption[Predicted affordance heatmap visualization]{
RGB observations and predicted affordance heatmaps are shown across representative manipulation skills.
}
\label{fig:heatmap_visualization}
\end{figure}

\paragraph{Affordance heatmap analysis.}
Figure~\ref{fig:heatmap_visualization} visualizes predicted heatmaps for representative simulation and real-robot skills.
The responses concentrate on regions where future interaction is expected, including object contacts, target containers, insertion areas, and stacking interfaces.
For motion-centric operations, the heatmaps extend along likely object-motion directions rather than only highlighting the current contact point.
These patterns suggest that the affordance heatmap branch captures trajectory-related contact and motion structure, providing explicit grounding for the object-interaction semantics used in action decoding.

\FloatBarrier
\section{Limitations}
1) Execution-level correction. MAST-VLA does not explicitly correct low-level deviations between predicted actions and realized end-effector motion. In longer or less controlled deployments, closed-loop feedback, temporal memory, and replanning may be needed to compensate for such deviations. 2) Depth-based spatial grounding. The trajectory-affordance prior assumes reasonably reliable RGB-D geometry. Although preprocessing handles routine depth noise and missing pixels, extreme RGB-D failure modes, such as heavy occlusion, reflective or transparent surfaces, and large calibration errors, can still degrade the spatial prior. Future work will explore recovery-aware execution, temporally aggregated or multi-view geometry, and broader skill taxonomies.
\section{Conclusion}
We presented MAST-VLA, a dual-conditioning VLA framework for multi-skill robot manipulation.
MAST-VLA couples skill-conditioned LoRA-MoE specialization in the LLM backbone with trajectory-affordance action guidance in the action head, enabling a shared policy to jointly model primitive-skill semantics and object-interaction semantics.
Experiments on CALVIN ABC$\rightarrow$D and real-robot sequence tasks show improved long-horizon execution stability, state-of-the-art simulation performance, and complementary gains from both conditioning paths.

\clearpage
\acknowledgments{
This work is supported by the National Natural Science Foundation of China (Grant No. 62502159, W2521174, 62476090, U23A20343, 62302167, 62222602), Fundamental Research Funds for the Central Universities, Natural Science Foundation of Shanghai (Grant No. 25ZR1402135, 25511103300, 25511104302), ``Chen Guang'' project supported by Shanghai Municipal Education Commission and Shanghai Education Development Foundation (Grant No. 25CGA22), Young Elite Scientists Sponsorship Program by CAST (Grant No. YESS20240780), Open Project Program of the State Key Laboratory of CAD\&CG (Grant No. A2501), Zhejiang University, Open Research Fund of Key Laboratory of Advanced Theory and Application in Statistics and Data Science-MOE, ECNU.
We thank VLA-Adapter and CALVIN teams for their open-sourced code and benchmark resources. We also thank the anonymous reviewers for their constructive feedback.
}

\bibliography{example,MASTVLA_added_references}  %

\clearpage
\appendix

\section{Additional Method Details}
\label{app:method_details}

\subsection{Primitive Skill Conditioning}
\label{app:skill_mapping}

MAST-VLA uses a coarse primitive skill condition as a lightweight signal for skill-aware adaptation. This condition is used only to retrieve learnable skill-query tokens and skill-dependent routing-bias vectors; it is not inserted into the language instruction as an additional text token and is not used as an action target.

For CALVIN, $s_t$ is assigned by matching each language instruction to benchmark-provided annotation variants using normalized lexical and fuzzy matching. The matched subtask is then grouped into a compact primitive skill taxonomy, which defines the indices used to retrieve the skill-query and routing-bias banks. This preprocessing is fixed across training and evaluation, providing a reproducible conditioning signal without modifying the original evaluation instructions.
The conditioning interface itself is agnostic to how primitive labels are obtained.
For broader task collections, the taxonomy can be constructed by clustering instruction/task representations or by using an LLM to partition and label the instruction corpus, followed by an instruction-to-skill classifier that provides the same conditioning indices.

\begin{table}[htbp]
\centering
\caption{Primitive skill taxonomy and skill-condition ids used by MAST-VLA.}
\label{tab:app_skill_mapping}
\small
\setlength{\tabcolsep}{4pt}
\begin{tabular}{p{0.12\linewidth}p{0.23\linewidth}p{0.55\linewidth}}
\toprule
\textbf{Skill ID} & \textbf{Primitive skill condition} & \textbf{Representative fine-grained commands} \\
\midrule
0 & Rotation & Rotate red/blue/pink blocks left or right. \\
1 & Sliding / pushing & Push red/blue/pink blocks left or right; push tabletop objects to target regions. \\
2 & Articulated-object operation & Move the slider left or right; open or close the drawer; close articulated containers. \\
3 & Lifting / picking & Lift red/blue/pink blocks to the table, slider, or drawer; pick tabletop objects. \\
4 & Placing / storing & Place objects onto the slider, into the drawer/box, or in other target regions. \\
5 & Stacking / nesting & Stack or unstack blocks/erasers; nest disposable cups in real-robot sequences. \\
6 & Switching & Turn the lightbulb or LED on/off. \\
\bottomrule
\end{tabular}
\end{table}

For real-robot data, each primitive instruction is associated with an operation type under the same primitive skill taxonomy.
During evaluation, the condition is updated at each primitive step rather than assigned once for the complete length-3 episode, allowing the same policy to switch primitive-skill conditions across consecutive subtasks while keeping a single shared VLA backbone.

\subsection{Skill-Conditioned LoRA-MoE Configuration}
\label{app:lora_moe_config}

The VLA backbone is based on a Qwen2.5-0.5B language model with a DINO/SigLIP 224-pixel fused vision backbone. We attach LoRA-MoE adapters to the selected MLP/FFN sublayers of the LLM backbone while keeping the original dense backbone weights frozen. Each LoRA-MoE adapter contains multiple low-rank expert updates, and the router selects top-$k$ expert updates for each token. In effect, the pretrained dense path remains unchanged, while MAST-VLA learns a sparse, skill-conditioned low-rank residual adaptation path.

The router computes content-dependent logits and adds a learnable skill routing bias before top-$k$ expert selection:
\begin{equation}
\label{eq:app_router_logits}
\begin{aligned}
\mathbf{z}_{i}
&=
\mathbf{W}_{\mathrm{gate}}\mathbf{h}_{i}
+
\mathbf{e}^{(\ell)}_{\mathrm{bias}}(s),\\
\mathcal{T}_{i}
&=
\operatorname{TopK}(\operatorname{softmax}(\mathbf{z}_{i}), k).
\end{aligned}
\end{equation}
Thus, skill conditioning shifts the routing preference but does not hard-assign a skill to a fixed expert. The selected experts are weighted by their normalized top-$k$ probabilities.

\begin{table}[htbp]
\centering
\caption{Main implementation settings of MAST-VLA.}
\label{tab:app_model_config}
\small
\setlength{\tabcolsep}{5pt}
\begin{tabular}{p{0.36\linewidth}p{0.52\linewidth}}
\toprule
\textbf{Item} & \textbf{Setting} \\
\midrule
Base policy & VLA-Adapter-Pro initialized from \texttt{prism-qwen25-extra-dinosiglip-224px-0\_5b}. \\
Language backbone & Qwen2.5-0.5B, hidden dimension 896. \\
Visual input & Static third-person RGB and wrist/secondary RGB, resized to $224\times224$. \\
Visual tokens & 256 patch tokens per image; 512 visual tokens for two-view input. \\
Action representation & Continuous action chunk with horizon $H=8$; each 7-D action contains 6-DoF end-effector motion and one gripper command. \\
Action query tokens & 64 learnable action tokens. \\
Primitive skill taxonomy & 7 coarse primitive skill conditions. \\
Skill query tokens & 32 learnable tokens per skill. \\
LoRA-MoE placement & LLM MLP/FFN sublayers. \\
Number of experts & 6 low-rank experts per LoRA-MoE adapter. \\
Routing & Content router plus skill routing bias; top-2 routing. \\
LoRA rank / scaling & Rank 64, scaling $\alpha=128$, dropout 0.0. \\
Router losses & Load-balancing coefficient $10^{-2}$ and router z-loss coefficient $10^{-4}$. \\
Trainable modules & LoRA expert updates, router/gate-skill-bias parameters, action queries, skill queries, proprio projector, action head, and affordance-heatmap head. \\
Frozen modules & Pretrained dense VLA backbone weights outside adapters, routers, query tokens, and heads. \\
\bottomrule
\end{tabular}
\end{table}

\subsection{Affordance Heatmap Head Architecture}
\label{app:heatmap_head}

The affordance-heatmap branch is implemented as a lightweight geometric prediction head rather than as a separate detector. Its inputs are the multi-layer VLA hidden states used by the action head and the static-view point map $\mathbf{P}\in\mathbb{R}^{H\times W\times 6}$, where each pixel stores world-coordinate xyz and normalized RGB.

First, a learnable query performs attention pooling over the selected VLA hidden states to obtain a global manipulation context vector. When proprioception is enabled, the current proprioceptive state is projected into the same hidden dimension and added to this context. Second, the point map is processed by a small CNN consisting of a $3\times3$ input convolution, two dilated $3\times3$ convolution blocks, and a $1\times1$ output convolution. The pooled VLA context generates FiLM parameters that modulate the spatial feature map before the final heatmap logits are predicted. This design lets the heatmap head combine language-conditioned VLA context with metric scene geometry.

The predicted affordance-heatmap logits are also used by the action head. We apply a sigmoid, resize the static-view heatmap to the patch grid, flatten it into a patch-level prior, and concatenate a neutral all-one prior for the wrist/secondary view. In the Pro action head, this prior is added as a soft attention bias for visual task tokens inside the residual cross-attention blocks. The bias increases the relative attention to interaction-relevant static-view patches while preserving access to the full visual context. During training, future gripper trajectories are used only to construct heatmap supervision. During inference, future trajectories are unavailable, and the action head uses the heatmap predicted from the current observation and instruction.

\section{Heatmap Construction and Training Details}
\label{app:training_details}

\subsection{Trajectory-Affordance Target Construction}
\label{app:heatmap_target_details}

The trajectory-affordance target is generated automatically from RGB-D observations and the future end-effector trajectory in the demonstration. In CALVIN, the static camera depth is dense after preprocessing, and we use the static RGB-D frame to construct a $200\times200$ xyz-rgb point map in the world frame.

For real-robot data, we apply the same geometric heatmap pipeline to calibrated RGB-D observations from the static RealSense camera. The depth frame is resized to $200\times200$ and converted to metric depth; pixels are marked valid only if they are finite and above a small positive depth threshold. To provide a dense point-map input to the heatmap head, invalid depth holes are filled by iterative neighbor averaging, remaining holes fall back to the median valid depth, and inpainted pixels are locally smoothed. We retain the original validity mask throughout this process. Camera intrinsics are scaled to the resized depth image, and the calibrated eye-to-hand extrinsic transform maps OpenCV camera-frame points into the robot/world frame. The final real-robot point map stores world-coordinate xyz and RGB at each pixel; originally invalid or inpainted pixels are tracked by the validity mask and down-weighted when computing the heatmap loss.

For each training sample, future end-effector positions are unnormalized from the action/proprioception statistics and compared with all visible surface points. The heatmap supervision target is defined as a Gaussian response to the nearest future end-effector position, with $\sigma=0.05$m and truncation radius $r_{\max}=0.20$m:
\begin{equation}
\label{eq:app_heatmap_target}
H^*(u)
=
\exp\left(
-\frac{\min_\tau \|\mathbf{x}_u-\mathbf{g}_\tau\|_2^2}{2\sigma^2}
\right)
\mathbb{I}\left[
\min_\tau \|\mathbf{x}_u-\mathbf{g}_\tau\|_2 \le r_{\max}
\right].
\end{equation}
This target highlights visible 3D scene surfaces that are close to the future end-effector trajectory and is the only heatmap target used for supervision.

The heatmap loss is a weighted binary cross-entropy on logits. Positive pixels are defined by $H^*>0.1$, and the positive class weight is computed per batch and clipped to $[1,20]$ to avoid domination by background pixels. When a point-validity mask is available, as in the real-robot data, the per-pixel loss is additionally reweighted so that originally valid depth pixels dominate supervision while inpainted regions have reduced influence. The heatmap term is weighted by $\lambda_{\mathrm{hm}}=0.005$ in the total training objective.

\subsection{Optimization Hyperparameters}
\label{app:optimization_details}

\begin{table}[htbp]
\centering
\caption{Training hyperparameters for the main MAST-VLA run.}
\label{tab:app_training_hparams}
\small
\setlength{\tabcolsep}{5pt}
\begin{tabular}{p{0.36\linewidth}p{0.52\linewidth}}
\toprule
\textbf{Item} & \textbf{Setting} \\
\midrule
Training data & CALVIN ABC training split. \\
Training schedule & 50k MAST-VLA fine-tuning updates. \\
GPUs & 8 GPUs with distributed data parallel training. \\
Batch size & 16 per GPU, gradient accumulation 1. \\
Optimizer & AdamW with parameter groups. \\
LoRA learning rate & $2\times10^{-4}$. \\
MoE router learning rate & $2\times10^{-5}$. \\
Skill query learning rate & $1\times10^{-4}$. \\
Action query learning rate & $5\times10^{-5}$. \\
Action head learning rate & $5\times10^{-5}$. \\
Heatmap head learning rate & $5\times10^{-5}$. \\
Image augmentation & Enabled during training. \\
Action loss & L1 regression over the 8-step action chunk. \\
Heatmap loss & Weighted BCE with logits. \\
Heatmap loss weight & $\lambda_{\mathrm{hm}}=0.005$. \\
MoE auxiliary losses & Load-balancing loss coefficient $10^{-2}$; z-loss coefficient $10^{-4}$. \\
\bottomrule
\end{tabular}
\end{table}

\section{Additional Experimental Protocol Details}
\label{app:experimental_protocols}

\subsection[CALVIN ABC-D Evaluation]{CALVIN ABC$\rightarrow$D Evaluation}
\label{app:calvin_protocol}

We follow the standard CALVIN ABC$\rightarrow$D long-horizon evaluation protocol~\citep{calvin}. The policy is trained on scenes A, B, and C, and evaluated in the unseen scene D. Evaluation uses 1,000 predefined initial-state and subtask-sequence pairs, where each sequence contains five consecutive subtasks. For each subtask, the rollout terminates once the official CALVIN task oracle detects success or once the per-subtask horizon of 360 environment steps is reached. The next instruction is issued only after the current subtask is completed.

Because each CALVIN evaluation sequence contains five subtasks, let $c_i\in\{0,\ldots,5\}$ denote the number of consecutive subtasks completed in the $i$-th sequence. We report
\begin{equation}
\label{eq:calvin_metrics}
\begin{aligned}
\mathrm{Success@}k
&=
\frac{1}{N}\sum_{i=1}^{N}\mathbb{I}[c_i\ge k],\\
\mathrm{Avg.\ Len}
&=
\frac{1}{N}\sum_{i=1}^{N}c_i
=
\sum_{k=1}^{5}\mathrm{Success@}k,
\end{aligned}
\end{equation}
where success rates are written as proportions in the final equality.
The numbers reported in the main paper use this official sequence-level metric.
The ablation baseline in Table~\ref{tab:ablation} is evaluated under the same implementation setting as our variants.

\subsection{Real-Robot Evaluation Protocol}
\label{app:real_robot_protocol}

Real-robot experiments are conducted on a Franka Research 3 arm with a Franka gripper and two Intel RealSense D435i RGB-D cameras. The input stream follows the same two-view design as simulation: a static third-person view for global scene layout and a wrist/secondary view for local manipulation feedback. Camera intrinsics and extrinsics are calibrated before data collection so that static-view depth can be represented in the robot/world frame.
Training and evaluation use the same two tabletop setups, object/task families, and primitive skill set, while object positions vary within approximately a $10\,\mathrm{cm}\times10\,\mathrm{cm}$ workspace region.
Language commands are paraphrased across demonstrations and evaluation, and the physical experiments naturally include illumination variation and RGB-D sensing noise.
No additional distractor objects are introduced in the reported evaluation.

Eight length-3 primitive sequences are evaluated across two tabletop scenes, with 20 trials per sequence for each method.
During each episode, one primitive instruction is issued at a time.
After the current primitive is completed, the scene is not reset and the next instruction is issued.
After a failure or after all three subtasks are completed, the episode terminates and the scene is reset for the next trial.
Human intervention is limited to resetting the scene between trials.

A pick is successful if the object is lifted above the table and remains in the gripper for a short holding period. A place/store primitive is successful if the object is released inside the target region or container. An insert primitive is successful if the flower is inserted into the target vase/container and remains there after release. A push/move primitive is successful if the object reaches the target region or intended displacement. A stack primitive is successful if the upper object remains stably on the lower object after release. A nest primitive is successful if one disposable cup is inserted into the other and remains nested after release.

\begin{table}[htbp]
\centering
\caption{Real-robot length-3 task sequences used in evaluation.}
\label{tab:app_real_robot_sequences}
\small
\setlength{\tabcolsep}{5pt}
\begin{tabular}{p{0.14\linewidth}p{0.10\linewidth}p{0.62\linewidth}}
\toprule
\textbf{Scene} & \textbf{Seq.} & \textbf{Primitive instruction sequence} \\
\midrule
Scene 1 & 1 & Pick cup $\rightarrow$ Nest cup $\rightarrow$ Push box \\
Scene 1 & 2 & Stack eraser $\rightarrow$ Push box $\rightarrow$ Pick cup \\
Scene 1 & 3 & Pick marker $\rightarrow$ Place marker $\rightarrow$ Stack eraser \\
Scene 1 & 4 & Push box $\rightarrow$ Pick marker $\rightarrow$ Place marker \\
Scene 2 & 1 & Pick flower $\rightarrow$ Insert flower $\rightarrow$ Stack blocks \\
Scene 2 & 2 & Stack blocks $\rightarrow$ Unstack blocks $\rightarrow$ Push mango \\
Scene 2 & 3 & Push mango $\rightarrow$ Pick flower $\rightarrow$ Insert flower \\
Scene 2 & 4 & Stack blocks $\rightarrow$ Push mango $\rightarrow$ Pick flower \\
\bottomrule
\end{tabular}
\end{table}

For each trial, let $c_i\in\{0,\ldots,3\}$ denote the number of consecutive primitive subtasks completed before the first failure. For length-3 sequences, we compute
\begin{equation}
\label{eq:real_robot_metrics}
\begin{aligned}
\mathrm{Success@}k
&=
\frac{1}{N}\sum_{i=1}^{N}\mathbb{I}[c_i\ge k],\\
\mathrm{Avg.\ Len}
&=
\frac{1}{N}\sum_{i=1}^{N}c_i
=
\sum_{k=1}^{3}\mathrm{Success@}k.
\end{aligned}
\end{equation}

\section{Additional Results and Failure Analysis}
\label{app:additional_results}

\subsection{Per-Sequence Real-Robot Results}
\label{app:real_robot_results}

\begin{table}[htbp]
\centering
\caption{Per-sequence real-robot evaluation results. Each method has 20 trials per sequence and 160 trials in total. Success rates are percentages and Avg. Len is out of 3.}
\label{tab:app_real_robot_results}
\small
\setlength{\tabcolsep}{5pt}
\begin{tabular}{lllcccc}
\toprule
\textbf{Scene} & \textbf{Seq.} & \textbf{Method} &
\multicolumn{3}{c}{\textbf{Tasks Completed in a Row (\%)}} &
\textbf{Avg. Len}$\uparrow$ \\
\cmidrule(lr){4-6}
& & & \textbf{1} & \textbf{2} & \textbf{3} & \\
\midrule

\multirow{20}{*}{Scene 1}
& \multirow{5}{*}{1}
& OpenVLA-OFT & 90 & 70 & 15 & 1.75 \\
& & $\pi_0$ & 100 & 85 & 50 & 2.35 \\
& & $\pi_{0.5}$ & 100 & 90 & 65 & 2.55 \\
& & VLA-Adapter-Pro & 95 & 85 & 30 & 2.10 \\
& & MAST-VLA & 100 & 85 & 85 & 2.70 \\

\cmidrule(l){2-7}
& \multirow{5}{*}{2}
& OpenVLA-OFT & 85 & 70 & 35 & 1.90 \\
& & $\pi_0$ & 95 & 80 & 40 & 2.15 \\
& & $\pi_{0.5}$ & 95 & 90 & 70 & 2.55 \\
& & VLA-Adapter-Pro & 90 & 75 & 30 & 1.95 \\
& & MAST-VLA & 95 & 95 & 90 & 2.80 \\

\cmidrule(l){2-7}
& \multirow{5}{*}{3}
& OpenVLA-OFT & 80 & 60 & 20 & 1.60 \\
& & $\pi_0$ & 100 & 90 & 35 & 2.25 \\
& & $\pi_{0.5}$ & 100 & 95 & 40 & 2.35 \\
& & VLA-Adapter-Pro & 95 & 80 & 45 & 2.20 \\
& & MAST-VLA & 100 & 95 & 60 & 2.55 \\

\cmidrule(l){2-7}
& \multirow{5}{*}{4}
& OpenVLA-OFT & 90 & 75 & 40 & 2.05 \\
& & $\pi_0$ & 90 & 80 & 35 & 2.05 \\
& & $\pi_{0.5}$ & 95 & 85 & 55 & 2.35 \\
& & VLA-Adapter-Pro & 85 & 75 & 30 & 1.90 \\
& & MAST-VLA & 100 & 90 & 75 & 2.65 \\

\midrule

\multirow{20}{*}{Scene 2}
& \multirow{5}{*}{1}
& OpenVLA-OFT & 85 & 75 & 20 & 1.80 \\
& & $\pi_0$ & 95 & 85 & 60 & 2.40 \\
& & $\pi_{0.5}$ & 100 & 90 & 65 & 2.55 \\
& & VLA-Adapter-Pro & 90 & 75 & 40 & 2.05 \\
& & MAST-VLA & 100 & 95 & 55 & 2.50 \\

\cmidrule(l){2-7}
& \multirow{5}{*}{2}
& OpenVLA-OFT & 95 & 75 & 45 & 2.15 \\
& & $\pi_0$ & 100 & 75 & 35 & 2.10 \\
& & $\pi_{0.5}$ & 100 & 80 & 50 & 2.30 \\
& & VLA-Adapter-Pro & 95 & 70 & 35 & 2.00 \\
& & MAST-VLA & 95 & 90 & 90 & 2.75 \\

\cmidrule(l){2-7}
& \multirow{5}{*}{3}
& OpenVLA-OFT & 80 & 65 & 25 & 1.70 \\
& & $\pi_0$ & 95 & 75 & 50 & 2.20 \\
& & $\pi_{0.5}$ & 100 & 85 & 65 & 2.50 \\
& & VLA-Adapter-Pro & 95 & 85 & 35 & 2.15 \\
& & MAST-VLA & 100 & 95 & 65 & 2.60 \\

\cmidrule(l){2-7}
& \multirow{5}{*}{4}
& OpenVLA-OFT & 90 & 65 & 30 & 1.85 \\
& & $\pi_0$ & 90 & 75 & 30 & 1.95 \\
& & $\pi_{0.5}$ & 95 & 80 & 35 & 2.10 \\
& & VLA-Adapter-Pro & 85 & 70 & 25 & 1.80 \\
& & MAST-VLA & 100 & 95 & 45 & 2.40 \\

\midrule

\multirow{5}{*}{Overall}
& \multirow{5}{*}{--}
& OpenVLA-OFT & 86.9 & 69.4 & 28.7 & 1.85 \\
& & $\pi_0$ & 95.6 & 80.6 & 41.9 & 2.18 \\
& & $\pi_{0.5}$ & 98.1 & 86.9 & 55.6 & 2.41 \\
& & VLA-Adapter-Pro & 91.2 & 76.9 & 33.8 & 2.02 \\
& & MAST-VLA & 98.8 & 92.5 & 70.6 & 2.62 \\

\bottomrule
\end{tabular}
\end{table}

Across all 160 real-robot trials, MAST-VLA reaches 98.8\% Success@1, 92.5\% Success@2, 70.6\% Success@3, and an Avg. Len of 2.62. Compared with OpenVLA-OFT, $\pi_0$, $\pi_{0.5}$, and VLA-Adapter-Pro, MAST-VLA improves Avg. Len by 0.77, 0.44, 0.21, and 0.60, respectively, indicating improved completion of full length-3 sequences under accumulated contact and pose errors.

\subsection{Failure Cases}
\label{app:failure_cases}

We observe three recurring failure modes. First, severe occlusion or noisy depth can produce incomplete trajectory-affordance targets, especially when the target surface is hidden by the gripper or by another object. Second, some primitive boundaries are semantically close, such as placing versus pushing into a container, and ambiguous language can lead to less stable skill conditioning. Third, long-horizon real-robot execution can accumulate small pose errors; a grasp or placement that is acceptable for the current primitive may leave the scene in a state that makes the next primitive harder. These cases are consistent with the limitations discussed in the main paper and motivate future work on latent skill inference and uncertainty-aware spatial priors.

\end{document}